\pdfoutput=1

\documentclass[11pt]{article}

\usepackage[]{acl}
\usepackage{tcolorbox}
\usepackage{times}
\usepackage{latexsym}
\usepackage{booktabs}
\usepackage{graphicx}
\usepackage{amsmath}
\usepackage{multirow}
\usepackage{pifont}
\usepackage{xspace}
\usepackage{amssymb}
\usepackage{hyperref}
\usepackage{natbib}

\usepackage[T1]{fontenc}

\usepackage[utf8]{inputenc}

\usepackage{microtype}
\usepackage{xcolor}

\newcommand{\model}{\textsc{Giraffe}-QwenVL\xspace} 
\newcommand{\qtwo}{\textsc{Giraffe}\xspace} 

\title{\qtwo: Design Choices for Extending the Context Length of Visual Language Models}

\author{Mukai Li$^1$\quad Lei Li$^1$\quad  Shansan Gong$^1$ \quad Qi Liu$^1$\\\
$^1$The University of Hong Kong \quad \\
  \texttt{kaikiaia3@gmail.com}
  \quad \texttt{ liuqi@cs.hku.hk}
}

\begin{document}
\maketitle
\begin{abstract}

Visual Language Models (VLMs) demonstrate impressive capabilities in processing multimodal inputs, yet applications such as visual agents, which require handling multiple images and high-resolution videos, demand enhanced long-range modeling. Moreover, existing open-source VLMs lack systematic exploration into extending their context length, and commercial models often provide limited details.
 To tackle this, we aim to establish an effective solution that enhances long context performance of VLMs while preserving their capacities in short context scenarios. Towards this goal, we make the best design choice through extensive experiment settings from data curation to context window extending and utilizing: (1) we analyze data sources and length distributions to construct ETVLM - a data recipe to balance the performance across scenarios; (2) we examine existing position extending methods, identify their limitations and propose M-RoPE++ as an enhanced approach; we also choose to solely instruction-tune the backbone with mixed-source data; (3) we discuss how to better utilize extended context windows and propose hybrid-resolution training. Built on the Qwen-VL series model, we propose \qtwo, which is effectively extended to 128K lengths. Evaluated on extensive long context VLM benchmarks such as VideoMME and Viusal Haystacks, our \qtwo achieves state-of-the-art performance among similarly sized open-source long VLMs and is competitive with commercial model GPT-4V.\footnote{\url{https://github.com/kiaia/GIRAFFE}}
\end{abstract}

\section{Introduction}
Visual Language Models (VLMs)~\citep{gpt4v,geminiteam2024geminifamilyhighlycapable} integrate visual and textual information, which are pivotal in understanding the multimodal world and excel in various applications, such as visual question answering and video understanding~\citep{liu2023llava,li2022blip}. 
However, more advanced scenarios involve multi-image and long video comprehension, which challenge the long-range modeling capabilities of VLMs. For instance, a 2K context length can only digest less than  a few frames~\citep{liu2023llava,liu2023llava15,li2023blip2}, limiting the upper bound of long video understanding.
Consequently, there is a pressing need for methods to extend the context window of VLMs and improve their performance in long context scenarios. This would benefit next-generation VLMs in performing long history visual agents or serving as world models~\citep{liu2023world}.  




Recent efforts for longer context VLMs focus on extending base Large Language Models (LLMs), along with visual alignment or efficient architectures. LongVA~\citep{zhang2024longva} seeks to transfer long context ability from language models to vision by modifying position embeddings in the LLM backbone (PI,~\citealt{chen2023extendingcontextwindowlarge}; NTK,~\citealt{Locallama}). LongVILA~\citep{xue2024longvilascalinglongcontextvisual} and LongLLaVA~\citep{wang2024longllavascalingmultimodalllms} accommodate longer sequences using multi-stage alignment and
instruction tuning~\citep{peng2023yarnefficientcontextwindow,fu2024data} with additional infrastructure and architecture. 
Despite these initial explorations, they have not investigated the feasibility of directly extending the context window of existing VLMs or systematically explored the design space in the extending pipeline. To bridge this gap, we decompose the challenge of extending context windows of existing VLMs into three fundamental research questions: \textbf{\textit{(1) How to effectively organize and curate training data? (2) How to efficiently train longer VLMs? (3) How to leverage the extended context window?}}


In our work, our goal is to answer the three research questions and find a solution in practice. 
To validate our design choices, 
we implement thorough experiments based on Qwen-VL series model~\citep{bai2023qwenvl, wang2024qwen2vlenhancingvisionlanguagemodels} and conduct comprehensive evaluations on single image understanding, image interleave,  and video tasks (\S\ref{subsec:tasks-construction}). 
For data curation, we prepare a diverse dataset comprising long context instruction data, multimodal instruction data, multimodal interleave data, and video instruction data (\S\ref{subsec:data-engineering}). We analyze the impact of different data compositions, ratios, and lengths on model performance (\S\ref{subsec:data-recipes}) and find that (1) short multimodal instruction data is crucial for both extending long context capability and retaining short context performance; (2) a balanced data ratio contributes to balanced performance on downstream tasks. 
For the second research question on extending training, we examine the effective context length of previous position embedding extending alternatives such as PI and NTK, discovering that, akin to LLM studies~\citep{gao2024trainlongcontextlanguagemodels, chen2024falls_short}, the effective length is shorter than the training length~(\S\ref{subsec:effective_length}). We propose M-RoPE++ (\S\ref{subsec:m-rope-ntk}) to extend position embedding on spatial and temporal dimensions. Validation experiments reveal that our method achieves better downstream task performance and longer effective length under the same training length (\S\ref{subsec:training-setting}). Different from LongVA~\citep{zhang2024longva} that first extend LLM base or LongLLaVA~\citep{wang2024longllavascalingmultimodalllms} and LongVILA~\citep{xue2024longvilascalinglongcontextvisual} that adopt multi-stage training with visual alignment and instruction tuning, we find that directly training VLMs by only updating LLM backbone's parameters achieves optimal results (\S\ref{subsec:multi-stage-training}).
To figure out how to use long context well in VLM, the third research question, we examine the trade-off between single-frame resolution and frame numbers regarding task performance (\S\ref{subsec:trade-off}). We consequently propose hybrid-resolution training, 
which further improves the utilization of a fixed context length (\S\ref{subsec:hybrid-inference}).

Based on our findings from the three research questions, we carefully select data recipes and training methods to extend Qwen-VL and Qwen2-VL to \model and \qtwo with 128K length. Our final models are evaluated on both short context tasks such as single image understanding and long context tasks with multi-image and long videos. Experimental results demonstrate that our \qtwo achieves state-of-the-art performance among long VLMs and there is a significant improvement for our \model compared with Qwen-VL base (\S\ref{sec:long-video-res}). Summarized contributions:

\begin{enumerate}
    \item We investigate different design choices to extend the context window of existing VLMs to 128K while maintaining comparable performance on short visual tasks. 
    \item Technically, M-RoPE++ and hybrid-resolution training methods are newly proposed by us to enhance model performance during training and inference.
    \item On existing long VLM benchmarks, \qtwo achieves state-of-the-art performance among similar scale open-sourced long VLMs and is competitive to commercial models.
\end{enumerate}


\section{How to Curate Extending Data}
\label{sec:q1}

Developing an effective recipe for extending the context window of VLMs is crucial. To systematically evaluate such recipes, we construct a comprehensive metric suite encompassing single-image, multi-image, and video tasks (\S\ref{subsec:tasks-construction}), enabling a thorough assessment of model performance across diverse scenarios. This section focuses on the selection and preprocessing of training data (\S\ref{subsec:data-engineering}), with an emphasis on understanding how data compositions, ratios, and lengths influence the model's capabilities (\S\ref{subsec:data-recipes}).

\subsection{Evaluation Tasks}
\label{subsec:tasks-construction}
We evaluate both long and short-context multimodal tasks, as it is essential for VLMs to sustain performance on short-context tasks after extended training. For short-context evaluation, we utilize widely adopted benchmarks such as single-image MME~\citep{fu2023mme} and MMBench~\citep{liu2024mmbenchmultimodalmodelallaround}, which capture the diverse capabilities of VLMs. For multi-image tasks, we incorporate Mantis-Eval~\citep{Jiang2024MANTISIM}, QBench~\citep{wu2024qbench}, and BLINK~\citep{fu2024blinkmultimodallargelanguage}, in line with LLaVA-Interleave~\citep{li2024llavanextinterleave}. Given the temporal nature of videos, which naturally represent long-context multimodal tasks, we evaluate on LongVideoBench~\citep{wu2024longvideobenchbenchmarklongcontextinterleaved} and VideoMME~\citep{fu2024videomme}. Additionally, we include the Visual Haystack Single Needle Challenge~\citep{wu2024visualhaystacks}, which requires locating specific visual information within a long sequence of images, providing a robust measure of the model's effective context length.

\subsection{Extending Data Curation}
\label{subsec:data-engineering}
\begin{table*}[!th]

    \centering
    \small
     \resizebox{\textwidth}{!}{
    \begin{tabular}{l|ll|l}
    \toprule
    
    Categories & Task types & Data sources & \%Part   \\
    \midrule
    \multirow{1}{*}{Text } & \multirow{1}{*}{Long context instructions} & LongAlign~\citep{bai2024longalignrecipelongcontext}, LongAlpaca~\citep{long-alpaca}  & \multirow{1}{*}{20\%}  \\
     \midrule
    \multirow{2}{*}{Image} & \multirow{1}{*}{Short visual instruction data} & LLaVA-Instruct~\citep{liu2023llava}, M3IT~\citep{li2023m3it}   & \multirow{1}{*}{25\%} \\
     \cmidrule(l){2-4}
    & \multirow{1}{*}{Image interleave data} & MMDU~\citep{liu2024mmdu}, Mantis~\citep{Jiang2024MANTISIM}, ArXivQA-interleave*   & \multirow{1}{*}{25\%} \\
    
     \midrule
    \multirow{2}{*}{Video } & Video QA & ShareGPT4O~\citep{chen2024far}, MLVU~\citep{MLVU}, LLaVA-Video~\citep{zhang2024llavavideo} & \multirow{2}{*}{30\%}  \\
     & Video Summary & ShareGPT4V~\citep{chen2023sharegpt4v}  \\
    
    \bottomrule
    \end{tabular}
     }
    \caption{Overview of our ETVLM training dataset. This dataset encompasses a wide range of modalities and is concatenated to target context length. * indicates that we reconstruct this data by our own.}
    \label{tab:data_stat}
    
\end{table*}
\begin{figure*}[!t]
    \centering
\includegraphics[width=0.990\linewidth]{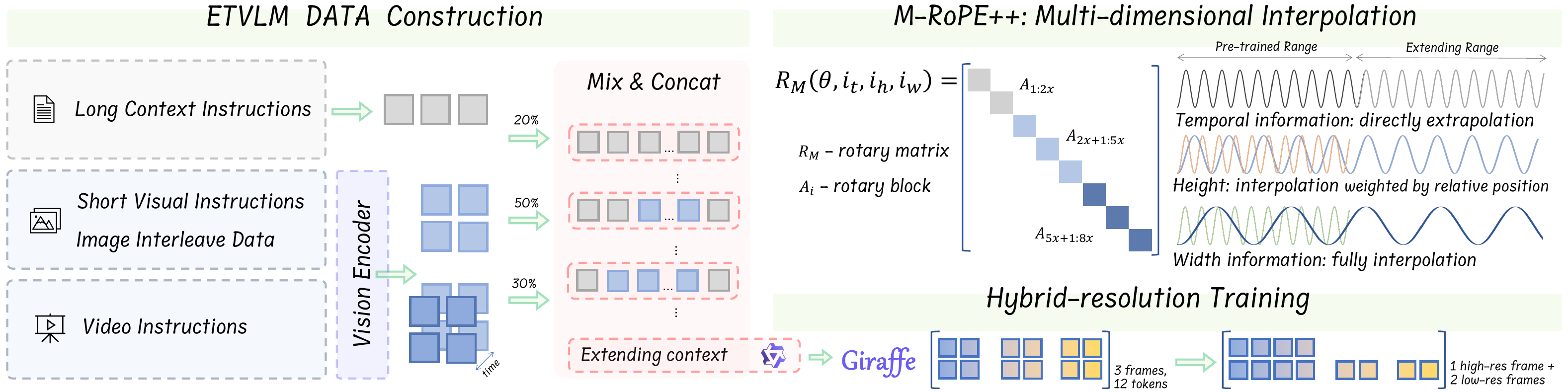}
    \caption{Pipeline of extending visual language models. We collect data from text, text-image pairs, and videos. We propose M-RoPE++ in extending training and hybrid-resolution inference to enhance the model performance.
    }
    \label{fig:pipeline}
\end{figure*}
\paragraph{Data Composition}


To construct our extended dataset, ETVLM, we incorporate four primary types of data with varying lengths: (i) Long-context instruction data, sourced primarily from LongAlign-10K~\citep{bai2024longalignrecipelongcontext} and LongAlpaca~\citep{long-alpaca}, with typical lengths ranging from 10K to 100K tokens. (ii) Short multimodal instruction data, drawn mainly from LLaVA-Instruct~\citep{liu2023llava} and M3IT~\citep{li2023m3it}. While the original datasets are generally under 10K tokens, we concatenate samples to achieve lengths between 10K and 100K tokens. (iii) Interleaved multimodal pre-training data, comprising multiple images with typical lengths of 1K–10K tokens, sourced from MMDU~\citep{liu2024mmdu} and Mantis~\citep{Jiang2024MANTISIM}. We also process interleaved image data from arXiv following the arXivQA protocol~\citep{li-etal-2024-multimodal-arxiv}. (iv) Long multimodal instruction data, created by sampling multiple frames from video datasets, primarily sourced from ShareGPT4V~\citep{chen2023sharegpt4v} and ShareGPT4O~\citep{chen2024far}. To address the scarcity of long video instruction data, we sample videos longer than 5 minutes from MLVU~\citep{MLVU}, ensuring MLVU is excluded from our test set to maintain fair evaluation. The data composition details are summarized in Table~\ref{tab:data_stat}.

\paragraph{Data Processing}
All data are processed into a dialogue format consistent with ChatML style~\citep{chatml}. Data are maintained in their original length and as concatenated multi-turn dialogues. For original-length text instruction data, we filter out special tokens. For short visual instruction and interleaved data, we adjust formatting and remove unnecessary symbols. Video data are sampled at 2 fps to reduce computational overhead. During data concatenation, we aim to match the target context length (e.g., 32K, 128K) as closely as possible without truncating content, ensuring a balance between efficiency and context preservation.
\subsection{Data Recipe Exploration}
\label{subsec:data-recipes}


We investigate the impact of different data ratio and data length on downstream task performance and provide recommendations for optimal data recipes. Using the same number of training tokens across all datasets, we conduct experiments with Qwen-VL~\citep{bai2023qwenvl} as the base model. 

\begin{figure}[!h]
    \centering
\includegraphics[width=0.950\linewidth]{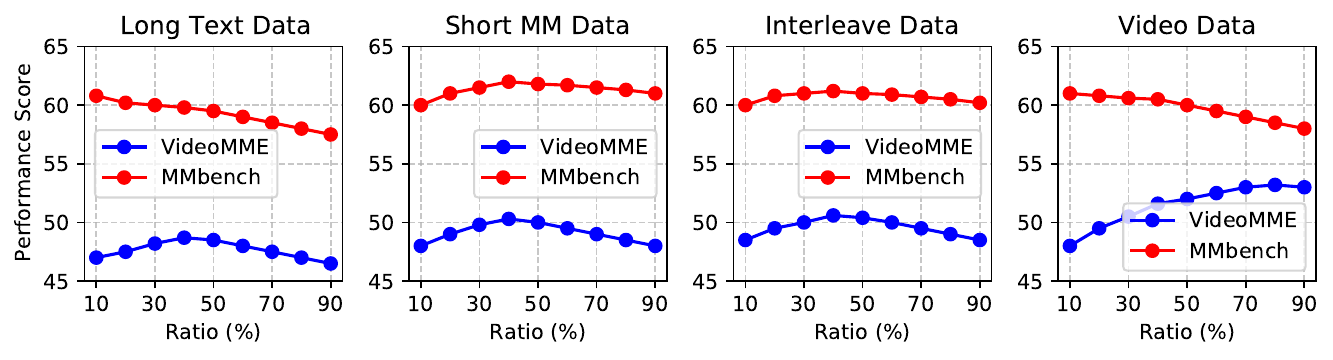}
    \caption{Performance of extending Qwen-VL with different data composition ratios.}
    \label{fig:data_ratio}
\end{figure}

\paragraph{Data Ratio}
To further investigate the impact of data composition on model performance, we conduct experiments by varying the proportion of a single data type from 10\% to 90\% while keeping the total training volume consistent. The results presented in Figure~\ref{fig:data_ratio} reveal that increasing the proportion of long video data improves long video comprehension but compromises performance on other tasks. Similarly, increasing the ratio of any specific data type predominantly enhances its associated downstream task performance. 
Based on these findings, we determine the final data composition strategy, as shown in Table~\ref{tab:data_stat}, which modestly increases the proportion of video data while reducing the share of pure text data. This adjusted recipe achieves a well-balanced performance across diverse task types.

\begin{figure}[!h]
    \centering
\includegraphics[width=0.950\linewidth]{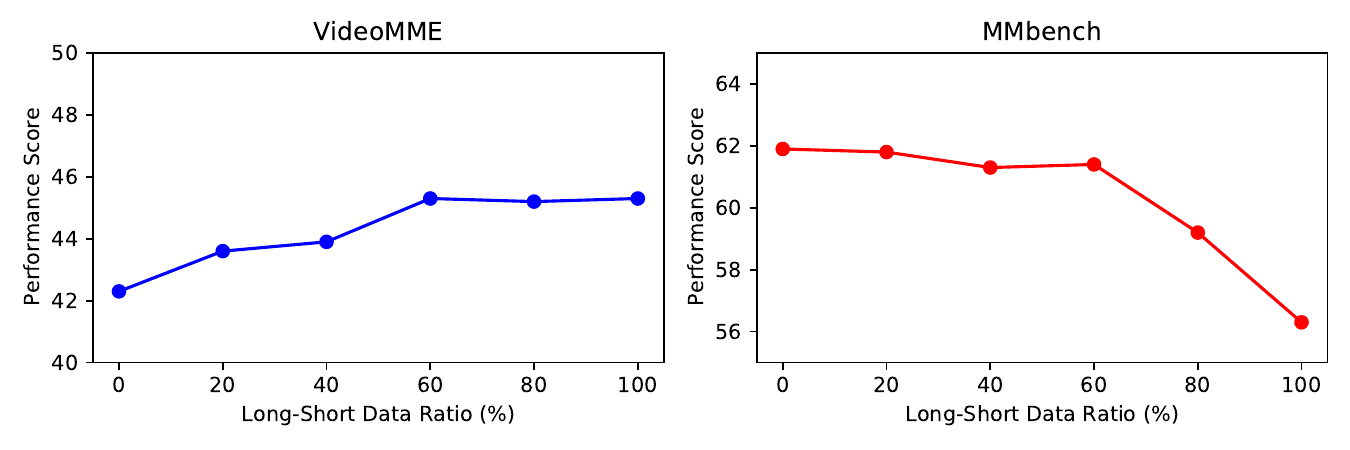}
    \caption{Performance on Qwen-VL trained with different composition ratio of long ($>$8K) and short data. }
    \label{fig:data_length}
\end{figure}

\paragraph{Data Length}
We categorize data into long data and short data based on whether their length exceeds 8K tokens. We investigate how different ratios of long and short data affect downstream performance on both long-context and short-context tasks. As shown in Figure~\ref{fig:data_length}, increasing the proportion of long data leads to improved performance on long-context tasks, with performance plateauing after the long data ratio reaches 60\%. However, for short-context tasks, when the proportion of long data exceeds 60\%, there is a notable decline in performance. This finding underscores the necessity of maintaining a sufficient amount of short data to preserve performance on short-context tasks. Based on these observations, we adopt a 60\% long data ratio for our extending training to achieve an optimal balance between long and short task performance.

\begin{tcolorbox}[colframe=blue!75!black, colback=blue!5, title=\textbf{Findings 1}] Short multimodal instruction data is crucial for both
extending long context capability and retaining short context performance. A balanced data ratio contributes to
balanced performance on downstream tasks. 
\end{tcolorbox}


\section{How to Extend Context Length}
\label{sec:q2}
In this section, we test the effective length of existing length-extending methods, address their limitations (\S\ref{subsec:effective_length}), and introduce our position embedding technique M-ROPE++ (\S\ref{subsec:m-rope-ntk}). We find that for extending VLMs, it is sufficient to tune the LLM base of VLMs without requiring multi-stage training (\S\ref{subsec:multi-stage-training}). We propose hybrid-resolution training to further leverage the fixed context length (\S\ref{subsec:hybrid-inference}). 

\subsection{Effective Length of VLMs}
\label{subsec:effective_length}
\begin{figure}[!h]
    \centering
    \includegraphics[width=0.990\linewidth]{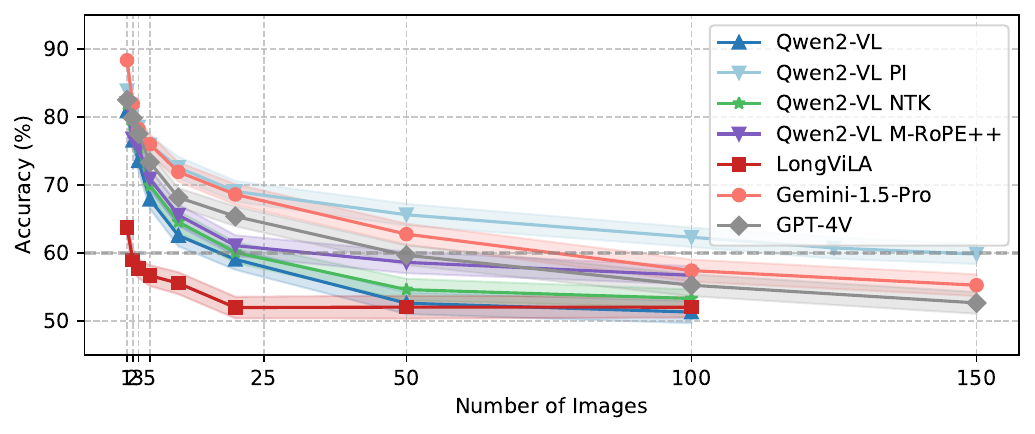}
    \caption{Results on visual haystack. The x-axis shows the number of input images, and the y-axis shows the retrieval success rate. The dashed line indicates the 60\% threshold for effective length.}
    \label{fig:effective_length}
\end{figure}
To evaluate the effective context length of VLMs, we draw inspiration from recent studies on LLMs, which suggest that their effective lengths are often only about half of their training lengths~\citep{chen2024falls_short, gao2024trainlongcontextlanguagemodels}. We adopt the single needle setting from Visual Haystack~\citep{wu2024visualhaystacks}, where models process varying numbers of input images and are tasked with identifying specific images and answering questions such as, "For the image with the anchor object, is there a target object?" This setup enables the assessment of performance across different context lengths, with random guessing yielding a 50\% success rate. All tests are conducted using native image resolutions consistent with the original configuration.

As shown in Figure~\ref{fig:effective_length}, retrieval success rates decrease as the number of input images grows. We define an accuracy threshold of 60\% to determine the effective length. The base Qwen2-VL model achieves effectiveness up to 15 images, corresponding to an effective length to approximately 10K tokens. After extending the training length to 128K tokens using existing length-extending methods like PI and NTK, the effective length increases to around 50 images, equivalent to approximately 40K tokens—still less than one-third of the training length. These findings highlight that the extended VLMs, similar to LLMs, exhibit the \textit{falls short} phenomenon~\citep{chen2024falls_short}, where effective length falls short of the training length. These findings highlight the need for a novel position-extending method to enhance the effective length of models.

\begin{tcolorbox}[colframe=blue!75!black, colback=blue!5, title=\textbf{Findings2}]The effective length in VLMs, including models that utilize existing position-extending methods, is smaller than the training length.
\end{tcolorbox}

\subsection{Position Extending on VLM}
In this subsection, we briefly introduce M-RoPE, discuss potential issues associated with existing position extending methods, and then present our proposed M-RoPE++ along with experimental results validating its effectiveness.
\label{subsec:m-rope-ntk}
\paragraph{M-RoPE}
Multimodal Rotary Position Embedding (M-RoPE) proposed in Qwen2-VL~\citep{wang2024qwen2vlenhancingvisionlanguagemodels} extends the RoPE~\citep{rope} to effectively model positional information with multi-dimensions. M-RoPE deconstructs the original rotary embedding into three components: temporal, height, and width.
The formal definition of M-RoPE and RoPE can be found in Appendix \ref{sec:appendix_rope}.

For a 16x-dimensional M-RoPE matrix, the dimensions are allocated in a 2:3:3 ratio for temporal, height, and width components respectively. This can be represented as:
\begin{equation}
\small
R_M(\theta, i_t, i_h, i_w) = 
\begin{bmatrix}
A_1 & 0 & \cdots & 0 \\
0 & A_2 & \cdots & 0 \\
\vdots & \vdots & \ddots & \vdots \\
0 & 0 & \cdots & A_{8x}
\end{bmatrix},
\end{equation}
where each $A_i \in \mathbb{R}^{2\times2}$ is a rotary block and $i_t$, $i_w$, $i_h$ are position indices. $\theta$ represents the rotary base. The blocks are allocated as follows:
\begin{itemize}
    \item $A_1$ to $A_{2x}$ represent the temporal dimension;
    \item $A_{2x+1}$ to $A_{5x}$ represent the height dimension;
    \item $A_{5x+1}$ to $A_{8x}$ represent the width dimension.
\end{itemize}

Each rotary block $A_i$ is defined as:
\begin{equation}
A_i = 
\begin{bmatrix}
\cos(i_x\theta_d) & -\sin(i_x\theta_d) \\
\sin(i_x\theta_d) & \cos(i_x\theta_d)
\end{bmatrix},
\end{equation}
where $i_x$ represents $i_t$, $i_h$, or $i_w$ depending on which dimension the block belongs to. The frequency basis $\theta$ is shared across all dimensions.

\paragraph{Position extending on M-RoPE}
In M-RoPE, the temporal index are allocated to the lower dimensions of the rotary embedding, which correspond to high-frequency information. Preserving this information is crucial for maintaining the model's ability to discern temporal order. Position extending methods such as position interpolation (PI;~\citealt{chen2023extendingcontextwindowlarge}) or modifying the RoPE base (NTK;~\citealt{Locallama}) tend to compress high-frequency signals indiscriminately, \textbf{potentially confusing the model's perception of order of close-by frames}. 
Conversely, the height and width dimensions occupy higher-dimensional spaces in the rotary embedding, indicating that they may not have fully covered the rotational domain during pre-training. This necessitates the application of interpolation to these dimensions. 
To address this, we propose M-RoPE++ that applies extrapolation exclusively to the temporal index and apply interpolation on height and weight index.

\paragraph{M-RoPE++}
We begin by defining key parameters following YaRN~(\citep{peng2023yarnefficientcontextwindow} :
\begin{equation}
s = \frac{L'}{L_V},
\end{equation}
where $s$ is the ratio between the extended context length $L'$ and the original visual context length $L_V$. 

We define $\lambda_d$ as the wavelength of the RoPE embedding at the $d$-th hidden dimension:
\begin{equation}
\lambda_d = \frac{2\pi}{\theta_d} = 2\pi b^{\frac{2d}{|D|}},
\end{equation}
and introduce the ratio $r$:
\begin{equation}
r = \frac{L'}{\lambda}.
\end{equation}

For M-RoPE, the index range is divided into three segments: temporal ($t$), height ($h$), and width ($w$). Temporal information is predominantly in high-frequency, which has been covered during pre-training stage. Therefore, we maintain extrapolation for this segment. For the height and width segments, where $\lambda > L'$, indicating insufficient rotational domain training, we employ interpolation to preserve their performance. This design is illustrated in Figure~\ref{fig:pipeline} right part.

We propose the following piecewise function to obtain the updated $\theta_d^{\prime}$ for M-RoPE++:
\begin{equation}
\small
\theta_d^{\prime} = 
\begin{cases}
\theta_d & \text{if } 0 < d \leq 2x, \\
(\frac{1}{s} + (1 - \frac{1}{s}) \cdot \frac{d - r_{5x}}{r_{2x} - r_{5x}}) \cdot {\theta_d} & \text{if } 2x < d \leq 5x, \\
\frac{\theta_d}{s} & \text{if } 5x < d \leq 8x.
\end{cases}
\end{equation}


\paragraph{Experiment Validation}
\label{subsec:training-setting}


We conduct a comparative analysis of various methods for extending the context length of VLMs, focusing on their performance on the VideoMME long context task and Single Needle Visual Haystacks in Table~\ref{tab:position_embedding_comparison}.

\begin{table}[t!]
    \centering
    \resizebox{0.48\textwidth}{!}{
    \begin{tabular}{lccccc|c}
    \toprule
    \textbf{Method} & \multicolumn{5}{c}{\textbf{VideoMME Long Score (Frames)}} &\textbf{VH(Images)} \\
    \cmidrule(lr){2-6} \cmidrule(lr){7-7}
     & 64 & 128 & 256 & 512 & 768 &100  \\
    \midrule
    Direct extrapolation & 52.5 & 54.3 & 56.0 & 55.4 & 55.6 &51.3\\
    PI training & 52.1 & 54.6 & 56.7 & 56.0 & 55.1  &57.8\\
    NTK-aware  &\textbf{53.8}  &54.8 &55.8 &56.2 &56.0 &56.7\\
    M-RoPE++ & 53.4 & \textbf{55.9} & \textbf{57.5} & \textbf{58.5} & \textbf{58.5}  &\textbf{61.3}\\
    \bottomrule
    \end{tabular}
    }
    \caption{Comparison of position embedding extension methods on VideoMME long video task and visual haystack on Qwen2-VL.
    }
    \label{tab:position_embedding_comparison}
    \end{table}

Our results demonstrate that M-RoPE++ consistently surpasses other methods, showing continued improvement as the number of frames increases in VideoMME Long tasks. This indicates that M-RoPE++ effectively captures long-range dependencies in video data. While direct extrapolation shows some potential for context extension, increasing the frame count without additional training does not lead to further performance gains. The PI method, due to significant interpolation of high-frequency information, exhibits slight performance degradation on shorter tasks. The NTK-aware approach achieves better results than the base model but still falls short of M-RoPE++ when handling higher frame counts, emphasizing the importance of preserving the original RoPE base in temporal dimensions. In the Visual Haystack test with 100 images, M-RoPE++ outperforms all baseline methods, demonstrating its ability to further enhance the effective length of VLMs. These findings highlight the effectiveness of M-RoPE++ in maintaining temporal order and extending context length in visual language models.

\begin{tcolorbox}[colframe=blue!75!black, colback=blue!5, title=\textbf{Findings 3}]The effective lengths achieved by existing position-extending methods remain insufficiently long. M-RoPE++ achieves better downstream task performance and longer effective length in the same training length. 
\end{tcolorbox}



\subsection{Multi-Stage Training}
\label{subsec:multi-stage-training}

\begin{table}[t!]
    \centering
    \resizebox{0.48\textwidth}{!}{
    \begin{tabular}{lccc}
    \toprule
    \textbf{Training Strategy} & \textbf{MMBench} & \textbf{BLINK} & \textbf{VideoMME} \\
    \midrule
    One-stage MM Instruction  & \textbf{82.8} & \textbf{54.6} & \textbf{58.5} \\
    Two-stage Text Extending + MM Instruction & 79.8 & 52.9 & 58.1 \\
    Two-stage MM Alignment + MM Instruction & 80.5 & 51.2 & 57.8 \\
    \bottomrule
    \end{tabular}
    }
    \caption{Comparison of different training strategies for extending Qwen2-VL context length. Results show that direct instruction tuning with long-context multimodal data performs best.}
    \label{tab:stage_comparison}
\end{table}


We investigate whether multi-stage training strategies commonly used in VLM training are necessary for extending context length. Previous works on long-context VLMs, typically training from an LLM base, often employ multiple stages, including extending the text-based model’s context length, multimodal alignment, and multimodal instruction tuning. For extending existing VLMs like Qwen2-VL, we explore two approaches: (1) extending the LLM base with additional pure text data (Wiki-103) followed by multimodal instruction data, like LongVA~\citep{zhang2024longva}, and (2) multimodal alignment using image-text pairs (Sampled from LAION-5B) followed by instruction tuning~\citep{xue2024longvilascalinglongcontextvisual, wang2024longllavascalingmultimodalllms}. As shown in Table~\ref{tab:stage_comparison}, our experiments indicate that pre-extending the text-based model with pure text data provides no significant advantage. This is likely because training with long-context multimodal data already addresses diverse length distributions, rendering pure text extension redundant. Moreover, performing multimodal alignment before instruction tuning degrades performance on short-context tasks. This could be attributed to Qwen2-VL already undergoing instruction tuning during pre-training; further tuning of MLP and ViT layers with alignment objectives may disrupt the model's learned distributions. With fixed training steps, this disruption negatively impacts short-context performance without yielding improvements for long-context multimodal tasks. These findings suggest that directly fine-tuning with long-context instruction data is the most effective approach for extending existing VLMs.

\begin{tcolorbox}[colframe=blue!75!black, colback=blue!5, title=\textbf{Findings 4}]Directly train VLM with mixed instruction data while only
updating LLM backbone’s parameters achieves optimal results.
\end{tcolorbox}


\subsection{Trade-off in Fixed Context Length}
\label{subsec:trade-off}
We explore the trade-off between the resolution of individual visual tokens for single images and the number of frames used, proposing a hybrid-resolution approach to improve long-context performance under a fixed context length. When encoding videos with a fixed total number of visual tokens, there exists an inherent balance between the resolution of each frame and the number of frames included. Prior work, such as Qwen2-VL~\citep{wang2024qwen2vlenhancingvisionlanguagemodels}, suggests that performance on image tasks tends to plateau at a certain resolution, beyond which additional visual tokens offer diminishing returns. To investigate this balance, we test various combinations of frame counts and resolutions, adjusting one in response to changes in the other. Table~\ref{tab:frame_resolution_tradeoff} summarizes the results of \qtwo on VideoMME medium and long sub-tasks under these configurations, highlighting the impact of different frame-resolution trade-offs.

\begin{table}[h]
\centering
\resizebox{0.48\textwidth}{!}{
\begin{tabular}{lccc}
\toprule
\textbf{Frame} & \textbf{Image Token} & \textbf{VideoMME} & \textbf{VideoMME} \\
\textbf{Count} & \textbf{Count} & \textbf{Medium} & \textbf{Long} \\
\midrule
128 & 960 & 62.5 & 55.6 \\
256 & 480 & 63.9 & 57.3 \\
512 & 240 & 64.6 & 58.2 \\
768 & 160 & \textbf{64.8} & \textbf{58.5} \\
768 & 120 & 64.3  &58.3 \\
1024 & 120 & 64.7 & \textbf{58.5} \\
\bottomrule
\end{tabular}
}
\caption{Performance of different frame counts and resolutions on VideoMME tasks for \qtwo.}
\label{tab:frame_resolution_tradeoff}
\end{table}

From the perspective of frame count, performance on medium-length tasks tends to plateau at 512 frames, with little to no substantial improvement beyond this threshold. For longer tasks, however, increasing the frame count continues to yield performance gains, despite a corresponding reduction in the resolution of each frame. Notably, when the frame count is high but individual frame resolution is already low, further compression of resolution negatively impacts performance. For medium-range tasks, a balanced approach combining a moderate frame count with higher per-frame resolution proves optimal. In contrast, long-range tasks benefit from higher frame counts to capture extended temporal dependencies, necessitating a careful trade-off in spatial resolution. These findings highlight the importance of a strategy that preserves high resolution for critical frames while accommodating longer sequences. To address this challenge, we propose hybrid-resolution training in \S\ref{subsec:hybrid-inference}, which provides an effective solution for optimizing performance in long-form video understanding tasks.

\subsection{Hybrid-resolution Training}
\label{subsec:hybrid-inference}
Video inference in VLMs often requires a substantial number of visual tokens, leading to significant computational overhead. A key challenge is to reduce token usage without notably compromising performance. Inspired by SlowFast~\citep{feichtenhofer2019slowfastnetworksvideorecognition}, we propose hybrid-resolution inference to address this issue. We partition the video frames into $N$ groups, each containing $L$ frames. For each group, we process the first frame using a high-resolution image that occupies $m$ visual tokens. The subsequent $L-1$ frames within the group are processed that occupy $\frac{m}{s}$ tokens, where $s$ is the compression ratio.
This approach significantly reduces the token usage from $L*N * m$ tokens to $ (1+\frac{L-1}{s})* N * m$ tokens. The high-resolution frames at the beginning of each group provide detailed visual information, while the low-resolution frames maintain temporal continuity and context at a reduced computational cost. This design is illustrated in Figure~\ref{fig:pipeline}.

\paragraph{Impact of hybrid-resolution training}
\begin{table}[h]
\centering
\resizebox{0.48\textwidth}{!}{
\begin{tabular}{lccccc}
\toprule
\textbf{Frames} & \textbf{(L,m,s)}   & \textbf{Avg. Image} &\textbf{VideoMME} & \textbf{VideoMME} \\
\textbf{Count} &  & \textbf{Tokens} &\textbf{Medium} & \textbf{Long} \\
\midrule
512 & (1,240,1) & 240 & 64.2 & 57.9 \\
512 & (4,240,3) & 120 & 64.0 & 57.6 \\
\midrule
1024 & (1,120,1) &120  & 64.7 & 58.5 \\
1024 & (4,240,3) &120 & \textbf{65.2} & \textbf{59.1} \\
\bottomrule
\end{tabular}
}
\caption{Performance comparison of hybrid-resolution training settings on VideoMME tasks.}
\label{tab:hybrid_res_inference}
\end{table}

The results in Table \ref{tab:hybrid_res_inference} demonstrate the effectiveness of hybrid-resolution training. Comparing the first two rows, we observe that reducing the resolution of low-res frames using hybrid-resolution inference only marginally affects downstream task performance while halving visual token usage. Furthermore, the bottom two rows reveal that under equivalent visual token constraints (i.e., 120 tokens per image), hybrid-resolution inference enables increased resolution for high-res frames and successfully enhances downstream task performance. These findings suggest that hybrid-resolution inference offers a promising approach to optimize the trade-off between computational efficiency and model performance in long-form video understanding tasks. We use (L,m,s)=(4,240,3) by default for other evaluations.

\begin{tcolorbox}[colframe=blue!75!black, colback=blue!5, title=\textbf{Findings 5}] Hybrid-resolution training can further
improve the utilization of a fixed context length.
\end{tcolorbox}

\section{Extended VLMs}
In this section, we first present the experimental setup and the relevant models, followed by an analysis of their performance across various downstream tasks.

\subsection{Infrastructure and Engineering}
\label{subsec:infrastructure}
We select Qwen2-VL, Qwen-VL as the base model for further training because of their strong performance at the 7B parameter scale and support for interleaved inputs. We employ the NTK method for Qwen-VL and M-RoPE++ for \qtwo to extend the model's window length. Training long VLMs results in substantial memory demands, thus we employ several optimization strategies to perform training on such long sequences. These
include FlashAttention-2~\cite{dao2022flashattention1,dao2023flashattention2},  Ring Attention~\cite{Liu2023RingAW}, ZERO~\cite{rajbhandari2020zero} (including activation checkpointing, and parameter
offload). To balance the load across 8 80G H100 GPUs, we shard the sequence in a zigzag way~\cite{zhu2023zigzag}. We use LoRA~\cite{hu2021lora} to reduce the GPU memory usage to train longer VLMs. We train the model for an average of 80 H100 hours.

\subsection{Models}
\begin{table*}[ht]
    \centering
    \resizebox{\textwidth}{!}{
    \begin{tabular}{l|ccccc|cccccc}
        \toprule
        \multirow{2}{*}{\textbf{Methods}} & \multirow{2}{*}{\textbf{Frames}} & \multicolumn{4}{c|}{\textbf{VideoMME}} & \multirow{2}{*}{\textbf{Frames}} &\multicolumn{4}{c}{\textbf{LongVideoBench}} & \multirow{2}{*}{\textbf{Avg}} \\
        \cmidrule(lr){3-6} \cmidrule(lr){8-11}
        & & \textbf{Short} & \textbf{Medium} & \textbf{Long} & \textbf{Overall} & &\textbf{(8, 15)} & \textbf{(15, 60)} & \textbf{(180, 600)} & \textbf{(900, 3600)} & \\
        \midrule
        \multicolumn{11}{c}{\textit{\centering Close-source VLMs}} \\
        \midrule
        GPT-4V (turbo) & 10  &70.5 &55.8 &53.5 &59.9 &256 &66.4 &71.1 &61.7 &54.5  &59.1 \\
        GPT-4o & 384  &80.0 &70.3 &65.3 &71.9 &256 &71.6 &\textbf{76.8} &\textbf{66.7} &\textbf{61.6}  &\textbf{66.7} \\
        Gemini-1.5-Pro &1/0.5fps &\textbf{81.7} &\textbf{74.3} &\textbf{67.4} &\textbf{75.0} &256 &\textbf{68.3} &73.2 &63.1 &56.3 &62.7 \\
        \midrule
        \multicolumn{11}{c}{\textit{\centering Open-source VLMs}} \\
        \midrule
        VideoLLaVA-7B & 8 & 45.3 & 38.0 & 36.2 & 39.9 & 8 & 43.1 & 44.6 & 36.4 & 34.4 & 39.1 \\
        VideoChat2-Mistral-7B & 16 & 48.3 & 37.0 & 33.2 & 39.5 &16 & 49.3 & 49.3 & 39.0 & 37.5 & 39.3 \\
        VideoLLaMA2-7B & 16 & 56.0 & 45.4 & 42.1 & 47.9 & -& - & - & - & - & - \\
        LLaVA-NeXT-Qwen2-7B & 32 & 58.0 & 47.0 & 43.4 & 49.5 & - & - & - & - & - & - \\
       
        LongVA-7B & 128 & 61.1 & 50.4 & 46.2 & 52.6 & - & - & - & - & - & -  \\
        LongVILA-8B & 256 & 61.8 & 49.7 & 39.7 & 50.5 & - & - & - & - & - & - \\

        \midrule
        Qwen-VL-Chat-7B & 4 & 46.9 & 38.7 & 37.8 & 41.1 & - & - & -& - & - & - \\
        \model & 128 & 55.4 & 51.2 & 46.9 & 51.2 & - & - & - & - & - & - \\
        Qwen2-VL-7B & 256 & \textbf{71.2} & 62.5 & 56.0 & 63.2 &256 & \textbf{67.8} & 70.4 & 56.6 & 51.3 & 61.5 \\
        \qtwo & 768 & 71.1 & 64.8  & 58.5& 64.8 &768 & 67.4 & 70.6 & 59.1 &55.9 & 63.3 \\

        \quad w/ Hybrid-res train\&inf & 1024 & 71.1 & \textbf{66.2} & \textbf{60.5} & \textbf{65.9} &1024 & 67.4 & \textbf{71.0} & \textbf{60.8} & \textbf{58.1} & \textbf{64.3} \\

        \bottomrule
    \end{tabular}
    }
    \caption{Performance comparison across VLMs on VideoMME and LongVideoBench tasks. We bold the best results for both close-source and open-source VLMs. We choose the best frames from our experiments in \S\ref{subsec:trade-off} and only use Hybrid-res inference on tasks above 512 frames. }
    \label{table:vlm_scores}
\end{table*}
We assess the following models:
\textbf{Qwen-VL-Chat-7B}~\cite{bai2023qwenvl} A visual language model based on the Qwen language model, incorporating visual capabilities through cross-attention and learnable query embeddings.
\textbf{VideoLLaVA-7B}~\cite{lin2024videollavalearningunitedvisual} A video-language model that extends LLaVA to handle video inputs, capable of processing up to 8 frames.
\textbf{VideoChat2-Mistral-7B}~\cite{li2024videochatchatcentricvideounderstanding} An advanced VLM built on the Mistral-7B, designed to process up to 16 frames.
\textbf{LongVA-7B}~\cite{zhang2024longva} A long context VLM based on Qwen-2 language model, utilizing a two-stage alignment process to handle up to 128 frames.
\textbf{LongVILA-8B}~\cite{xue2024longvilascalinglongcontextvisual} A long context VLM based on VILA language model, capable of processing up to 256 frames.
\textbf{Qwen2-VL}~\cite{wang2024qwen2vlenhancingvisionlanguagemodels} A foundational VLM that employs dynamic image tokenization and M-RoPE, with pre-trained 16K context length. We train and evaluate our \model and \qtwo in this section with the best setting shown in \S\ref{sec:q1} and \S\ref{sec:q2}.

\subsection{Video Task Results}
\label{sec:long-video-res}
Our extended models, \model and \qtwo, demonstrate substantial improvements in video understanding across various temporal scales while specifically maintaining competitive performance on short videos. Table~\ref{table:vlm_scores} shows that \model significantly outperforms its base model Qwen-VL-Chat, enabling better understanding of video content. Notably, \qtwo, based on an improved base model and capable of processing 1024 frames, achieves state-of-the-art performance among open-source models in both VideoMME and LongVideoBench, even surpassing GPT-4V in several categories. These results provide compelling evidence that our approach successfully extends the context window of VLMs, particularly benefiting long context video understanding tasks while reserving original short context capacities. 

\subsection{Multi Image Task Results}
\begin{table}[t!]
    \centering
    \resizebox{0.48\textwidth}{!}{
    \begin{tabular}{lcccc}
        \toprule
        \textbf{Model} & \textbf{Mantis-Eval} & \textbf{QBench} & \textbf{BLINK} \\
        \midrule
        LLaVA-v1.5-7B & 31.3& 49.3 & 37.1 \\
        GPT-4V &62.7 &76.5 & 51.1 \\
        Qwen-VL   &39.2 &45.9  &31.1  \\
        \model & 48.3 &57.4  & 41.2 \\
        Qwen2-VL & \underline{63.4} &\textbf{76.9}  & \underline{53.3} \\
        \qtwo & \textbf{63.9} & \underline{76.8} &\textbf{54.5} \\
        \bottomrule
    \end{tabular}
    }
    \caption{VLMs results on multi-image scenario: Mantis-Eval, QBench and BLINK. We bold the best results and underline the second best.}
    \label{table:interleave_comparison}
\end{table}
In the multi-image evaluation presented in Table~\ref{table:interleave_comparison}, \model exhibits substantial improvements, whereas \qtwo also demonstrates enhancements, validating the efficacy of our pipeline. 
In multi-image scenarios, context length is less critical than in long video tasks. Qwen-VL's superior performance stems from capacities trained on the ETVLM dataset, compared to its initial 2K context length. In contrast, Qwen2-VL has already undergone substantial pre-training in 16K contexts. Additionally, Qwen2-VL benefits from a broader range of training data compared to Qwen-VL, rendering the incremental advantages from ETVLM data relatively modest.

\subsection{Image Task Results}
\begin{table}[t!]
    \centering
    \resizebox{0.48\textwidth}{!}{
    \begin{tabular}{lcccc}
        \toprule
        \textbf{Model} & \textbf{MME$_{p}$} & \textbf{MME$_{c}$} & \textbf{MMBench$(en)$} \\
        \midrule
        GPT-4V & 1590.5  & 573.2 & 82.8 \\
        Qwen-VL & 1487.6 & 360.7 & 60.9 \\
        \model & 1489.7 & 372.9 & 61.5 \\
        Qwen2-VL & \textbf{1695.3}  & \underline{1630.4} & \textbf{82.8} \\
        \qtwo & \underline{1692.9} & \textbf{1635.4} & \underline{82.1} \\
        \bottomrule
    \end{tabular}
    }
    \caption{VLM performance on the single-image scenario: MME and MMBench tasks. We bold the best results and underline the second best.}
    \label{table:mme_mmbench_comparison}
\end{table}
The results from Table~\ref{table:mme_mmbench_comparison} demonstrate that our \qtwo maintains competitive performance on short-form multimodal tasks. This balanced capability can be attributed to our training strategy, which incorporates a mix of short instruction data alongside long context video inputs. Incorporating LLaVA-Instruct and M3IT in our training process ensures the model retains its capacity in single-image understanding.

\section{Related Work}
We list the related work for extending the existing LLMs and VLMs.
\subsection{Long Context Language Models}
With the rise of LLMs, research has focused on extending their capacity for longer contexts. The main solution involves addressing the out-of-distribution issue with position-embedding and enhancing model extrapolation capabilities. Training-free methods like streamingLLM~\citep{xiao2024efficient}, InfLLM~\citep{xiao2024infllm} and ChunkLLaMA~\citep{an2024trainingfree} offer cost-effective ways to scale context window size. Additionally, further training using modified RoPE~\citep{rope} base frequency is introduced in NTK~\citep{Locallama}, PI~\citep{chen2023extendingcontextwindowlarge} and YaRN~\citep{peng2023yarnefficientcontextwindow}, a effective practice adopted by models such as CodeLlama~\citep{rozière2024codellamaopenfoundation} and LLaMA 3.1~\citep{dubey2024llama3herdmodels}. Moreover, efforts have also been made on data curation for long context training~\citep{bai2024longalignrecipelongcontext, gao2024prolong, fu2024data}. However, corresponding comprehensive studies on extending context for open-source VLMs remain limited.
\subsection{Visual Language Models}
Advancements in LLMs are driving the evolution of VLMs. 
MiniGPT-4~\citep{zhu2023minigpt4} and LLaVA-series~\citep{liu2023llava,liu2023llava15,li2024llavanextinterleave,2023llavarlhf} integrate visual modalities into the LLM architecture using visual encoders.
The recent Qwen-VL series~\citep{bai2023qwenvl,wang2024qwen2vlenhancingvisionlanguagemodels} significantly enhances multimodal training pipeline, distinguishing itself with strong performance among open-source VLMs. VideoLLaVa~\citep{lin2024videollavalearningunitedvisual} and VideoChat~\citep{li2024videochatchatcentricvideounderstanding} focus on video scenarios requiring long context windows, yet their context windows are still limited.
For long context VLMs, recent LongVA~\citep{zhang2024longva} are first extending an LLM base model to 128K token lengths and then developing it into a VLM. Concurrent work LongVILA~\citep{xue2024longvilascalinglongcontextvisual} also involves multi-stage training starting from an LLM backbone and employs an improved sequence parallel technique for efficient training, while LongLLaVA~\citep{wang2024longllavascalingmultimodalllms} combines Mamba and Transformer blocks to reduce memory usage. In contrast, our model \qtwo optimizes various data recipes and position extending designs, establishing itself as the state-of-the-art among open-source long VLMs.

\section{Conclusion and Future Work}
We develop an effective solution to extend the context length of VLMs while preserving their performance on shorter contexts. Our comprehensive experiments led to the introduction of the ETVLM dataset for extended training and M-RoPE++ for improved position embedding learning. We use Hybrid-res training to better use long context window. Our extended model, \qtwo, achieves state-of-the-art performance for long context tasks. In the future, we aim to apply \qtwo to more complex scenarios, such as long-term history multi-modal chats and visual agents in real-world applications.

\bibliographystyle{acl_natbib}

\newpage
\appendix

\section{RoPE and M-RoPE}
\label{sec:appendix_rope}
Attention is defined over $C$ embeddings $X = [x_1, x_2, \ldots, x_C]^T \in \mathbb{R}^{C\times d}$ where $d$ is the model dimension. Learned weight matrices $W_v \in \mathbb{R}^{d\times d_k}$, $W_q \in \mathbb{R}^{d\times d_k}$, and $W_k \in \mathbb{R}^{d\times d_k}$ are used to transform these inputs where $d_k$ is the projected hidden dimension. The attention mechanism itself computes the attention matrix and applies it to produce a weighted sum of the value vectors:

\begin{equation}
\small
\text{Attention}(Q, K, V) = AV = \text{softmax}\left(\frac{QK^T}{\sqrt{d_k}}\right)V.
\end{equation}

Basic attention was originally defined with: $Q = XW_q$, $K = XW_k$, $V = XW_v$. However, this approach does not directly encode the relative position of keys and values.

Rotary Position Embeddings (RoPE) \cite{sun2022lengthextrapolatabletransformer} encode positional information by applying a phase rotation to each element of the embedding vectors. Formally, we define a transformation $f$:

\begin{equation}
f_W(x_i, \theta) = R(\theta, i)W^Tx_i
\end{equation}

Here $x_i \in \mathbb{R}^{d_k}$ is an embedding for position $i$, $W$ is a projection matrix, and $\theta \in \mathbb{R}^{d_k/2}$ is a frequency basis. The function is defined based on the rotary position matrix:

\begin{equation}
\scriptsize
R(\theta, i) = 
\begin{bmatrix}
\cos i\theta_1 & -\sin i\theta_1 & \cdot\!\cdot\!\cdot & 0 & 0 \\
\sin i\theta_1 & \cos i\theta_1 & \cdot\!\cdot\!\cdot & 0 & 0 \\
\vdots & & & & \vdots \\
0 & 0 & \cdot\!\cdot\!\cdot & \cos i\theta_{d_k/2} & -\sin i\theta_{d_k/2} \\
0 & 0 & \cdot\!\cdot\!\cdot & \sin i\theta_{d_k/2} & \cos i\theta_{d_k/2}
\end{bmatrix}
\end{equation}

Due to the arrangement of frequencies, this matrix has the property that $R(\theta, n - m) = R(\theta, m)^TR(\theta, n)$ by Ptolemy's identity. We redefine the query-key product between two positions $m$ and $n$ as,

\begin{equation}
q_m^Tk_n = f_{W_q}(x_m, \theta)^Tf_{W_k}(x_n, \theta)
\end{equation}

Multimodal Rotary Position Embedding (M-RoPE) extends the concept of RoPE to effectively model positional information of multimodal inputs. M-RoPE deconstructs the original rotary embedding into three components: temporal, height, and width. For text inputs, these components utilize identical position IDs, making M-RoPE functionally equivalent to 1D-RoPE. For image inputs, the temporal IDs remain constant, while distinct IDs are assigned to the height and width components based on the token's position in the image. For video inputs, the temporal ID increments for each frame, while the height and width components follow the same ID assignment pattern as images.

Formally, we define the M-RoPE transformation function $f_M$ as:

\begin{align}
f_M(x_i, \theta_t, \theta_w, \theta_h) = [&R_t(\theta_t, i_t)W_t^Tx_{it}; \notag \\
&R_w(\theta_w, i_w)W_w^Tx_{iw}; \\
&R_h(\theta_h, i_h)W_h^Tx_{ih}] \notag
\end{align}

where $x_i$ is the embedding vector, $\theta_t$, $\theta_w$, $\theta_h$ are frequency bases,
$i_t$, $i_w$, $i_h$ are position indices, and $W_t$, $W_w$, $W_h$ are projection matrices
for temporal, width, and height dimensions respectively.

The query-key product for M-RoPE is then redefined as:

\begin{equation}
\small
q_m^Tk_n = f_M(x_m, \theta_t, \theta_w, \theta_h)^Tf_M(x_n, \theta_t, \theta_w, \theta_h)
\end{equation}

For a 16x-dimensional M-RoPE matrix, the dimensions are allocated in a 2:3:3 ratio for temporal, height, and width components respectively. This can be represented as:
\begin{equation}
\small
R_M(\theta, i_t, i_h, i_w) = 
\begin{bmatrix}
A_1 & 0 & \cdots & 0 \\
0 & A_2 & \cdots & 0 \\
\vdots & \vdots & \ddots & \vdots \\
0 & 0 & \cdots & A_{8x}
\end{bmatrix}
\end{equation}

where each $A_i \in \mathbb{R}^{2\times2}$ is a rotary block. The blocks are allocated as follows:
\begin{itemize}
    \item $A_1$ to $A_{2x}$ represent the temporal dimension
    \item $A_{2x+1}$ to $A_{5x}$ represent the height dimension
    \item $A_{5x+1}$ to $A_{8x}$ represent the width dimension
\end{itemize}

Each rotary block $A_i$ is defined as:
\begin{equation}
A_i = 
\begin{bmatrix}
\cos(i_x\theta_d) & -\sin(i_x\theta_d) \\
\sin(i_x\theta_d) & \cos(i_x\theta_d)
\end{bmatrix}
\end{equation}

where $i_x$ represents $i_t$, $i_h$, or $i_w$ depending on which dimension the block belongs to. The frequency basis $\theta$ is shared across all dimensions.

This formulation allows M-RoPE to effectively model multimodal inputs while maintaining the rotary structure for each dimension.

\section{Impact of RoPE Base}
\label{sec:rope_base}

We investigated the effect of different RoPE bases on the performance of Qwen-VL. Our findings indicate that the optimal performance was achieved by following the recommendations from Su's blog, specifically using a RoPE base of 500,000 for a context length of 128k. Increasing the base beyond this point did not yield significant improvements while keeping the default base of 10,000 resulted in a notable performance drop. Table \ref{tab:rope_base_comparison} summarizes our results.

\begin{table}[h]
\centering
\resizebox{0.48\textwidth}{!}{
\begin{tabular}{lcccc}
\toprule
\textbf{RoPE Base} & \textbf{VideoMME Long} & \textbf{VideoMME Avg} & \textbf{MME Sum} & \textbf{MMBench} \\
\midrule
10,000 (default)  & 39.5  & 41.1 & 1848.29 & 60.9 \\
500,000 (optimal)  & \textbf{43.2}  & \textbf{51.2}& \textbf{1862.62} & \textbf{61.5} \\
1,000,000 & 43.1 & 51.1  & 1862.20 & 61.4 \\
\bottomrule
\end{tabular}
}
\caption{Performance comparison of different RoPE bases across various benchmarks.}
\label{tab:rope_base_comparison}
\end{table}

These results underscore the significance of meticulously adjusting the RoPE base when expanding the context window of visual language models. Our findings corroborate the conclusions presented in Su's blog \cite{kexuefm-10122}, which posits that for models with a context length of 128k, an optimal RoPE base of $4.9 \times 10^6$ is recommended. This value closely approximates our selected base of $5 \times 10^5$, which consistently demonstrates superior performance compared to the default configuration across all evaluated metrics.

Interestingly, further increasing the base beyond this point does not yield significant performance improvements. This observation is consistent with the approaches taken by models like LLaMA 2 and Qwen, which have opted for even larger base values. Such choices may provide additional flexibility for future extensions of model context lengths.

The effectiveness of the optimized RoPE base in capturing long-range dependencies in multimodal data underscores the critical role of position embedding strategies in enhancing the performance of extended visual language models.


\section{Progressive Extending}
\label{subsec:progressive-training}
To ensure more stable training, we adopted a progressive extending strategy. For \model, we set multiple incrementally increasing context lengths: 8K, 32K, 64K, and 128K. We concatenate and chunk ETVLM data according to these different context lengths. For \model, we investigate the optimal RoPE base setting, as detailed in Appendix~\ref{sec:rope_base}. Following~\citet{kexuefm-10122}, we experiment with bases of $5\times 10^4$, $1\times 10^6$, $2.5\times 10^6$, and $5\times 10^6$. For \qtwo, we employ M-RoPE++, training up to 64K before extending to 128K. This approach allows the model to gradually adapt to longer sequences while maintaining performance on shorter contexts.
\paragraph{Ablation of progressive extending}
We conduct comparative experiments on Qwen-VL to evaluate two methods for extending the model's context length: a single-stage approach and a progressive multi-stage approach. Both methods are using the same number of training steps. The results are summarized in Table~\ref{tab:extension_comparison}.
\begin{table}[t!]
\centering
\resizebox{0.48\textwidth}{!}{%
\begin{tabular}{lccc}
\toprule
\textbf{Method} & \textbf{MME$_P$} & \textbf{MME$_c$} & \textbf{VideoMME} \\
\midrule
Single-stage (2k$\rightarrow$128k) & 1462.58 & 350.71 & 48.9 \\
Progressive  & \textbf{1487.58} & \textbf{360.71} & \textbf{51.2} \\
\bottomrule
\end{tabular}%
}
\caption{Comparison of single-stage and progressive extension methods on Qwen-VL.}
\label{tab:extension_comparison}
\end{table}
Our experiments demonstrate that the progressive extending approach consistently outperforms the single-stage method across different evaluated tasks. This suggests that gradually increasing the context length during training allows the model to better adapt to longer sequences, resulting in improved performance on various tasks.

\end{document}